\documentclass[a4paper,12pt]{article}

\usepackage[ansinew]{inputenc}
\usepackage{natbib}
\usepackage{amsmath}    
\usepackage{amssymb}
\usepackage{graphicx}  
     
\def\argmax{\mathop{\rm argmax}}
\newcommand{\trademark}{\raisebox{1ex}{\scriptsize\textregistered}}

\begin{document}

\title{\LARGE Modeling and Control with Local Linearizing Nadaraya Watson Regression}
\date{}

\author{\small Steffen K\"uhn and Clemens G\"uhmann\\ \small Technische Universit\"at Berlin\\ \small Chair of Electronic Measurement and Diagnostic Technology\\ \small Einsteinufer 17, 10585 Berlin, Germany\\ \small \{steffen.kuehn, clemens.guehmann\}@tu-berlin.de}

\maketitle

\begin{abstract}
Black box models of technical systems are purely descriptive. They do not explain why a system works the way it does. Thus, black box models are insufficient for some problems. But there are numerous applications, for example, in control engineering, for which a black box model is absolutely sufficient. In this article, we describe a general stochastic framework with which such models can be built easily and fully automated by observation. Furthermore, we give a practical example and show how this framework can be used to model and control a motorcar powertrain.
\end{abstract}

\section{Introduction}

The modeling of technical systems is of eminent importance in engineering. On the one hand, models are used to simulate systems in order to detect errors during the design process. On the other hand models are developed because
\begin{itemize}
\item Tests with existing systems are too expensive or too dangerous,
\item Simulation results have to be reproducible or,
\item To detect a deviation of the normal behavior of a system.  
\end{itemize}
Another aspect is the control of technical systems, as we will demonstrate in section~\ref{Selbstlernender Regler}.

There are essentially two different ways to generate models. The first method is to let a specialist build the model manually, using his technical expertise. This is the most common method and unavoidable for systems that do not yet exist. The second way is to generate models by observation and description. When the internals are not essential for the technical problem, this method is an interesting option, especially for complicated systems. 

This article is dedicated to the second way, which is called \textit{system identification}. While numerous papers deal with this topic, the framework presented here is designed to have a much wider application range. In fact, it can be used for classification, recognition, regression, prediction, and reconstruction of disturbed patterns. This becomes possible due to a strict stochastic problem definition and a separation of data storing and data processing. For data storing, we use an online kernel density estimation approach with compression abilities and the possibility to forget unused knowledge. The data processing uses a modified version of the Nadaraya-Watson regression. A special capability is the possibility to change arbitrary input and output of the model and to mark some data as corrupt. This significantly increases the flexibility. A model and control strategy of a motorcar powertrain will demonstrate these possibilities of the new framework.

\section{Basics} 

\subsection{Nadaraya-Watson Regression}

Probability densities are ideal for the modeling of uncertain knowledge. Commonly, technical relations are described by functions. For example, the function $\boldsymbol{y} = \boldsymbol{f}(\boldsymbol{x})$ maps a value $\boldsymbol{x} \in \mathbb{R}^{d_x}$ to a value $\boldsymbol{y} \in \mathbb{R}^{d_y}$. The inherent assumption is that the function $\boldsymbol{f}$ and the value $\boldsymbol{x}$ are known with infinite certainty, which is mostly only approximately correct.

The second, more general option is to model the conditional probability density
$p_{\boldsymbol{X},\boldsymbol{Y}}(\boldsymbol{x},\boldsymbol{y})$. Because both parameters $\boldsymbol{x}$ and $\boldsymbol{y}$ can be interpreted as realizations of random variables, it is possible to describe 'knowledge' and 'uncertainty' in the same model. If a value $\boldsymbol{x}_0$ is given, the related most probable value $\boldsymbol{y}_0$ can be computed by
\begin{equation}
\boldsymbol{y}_0 = \argmax\limits_{\forall\boldsymbol{y}}\left\{p_{\boldsymbol{X},\boldsymbol{Y}}\left.(\boldsymbol{x},\boldsymbol{y})\right|_{\boldsymbol{x} = \boldsymbol{x}_0}\right\}. 
\label{DAGMControl_f10}
\end{equation}
This can also be applied for the opposite case, where $\boldsymbol{y}_0$ is given and we want to find the most probable value for $\boldsymbol{x}$. But this maximization is usually too costly. Although it seems to be very restrictive, for simple, unimodal probability densities, the expression
\begin{equation}
\boldsymbol{y}_0 \approx \int\limits_{\forall \boldsymbol{y}}\,\boldsymbol{y}\, p_{\boldsymbol{Y}|\boldsymbol{X}}(\boldsymbol{y}|\boldsymbol{x}_0) \mathrm{d}\boldsymbol{y} \label{DAGMControl_f1}
\end{equation}
with
\begin{equation}
p_{\boldsymbol{Y}|\boldsymbol{X}}(\boldsymbol{y}|\boldsymbol{x}) = \frac{p_{\boldsymbol{X},\boldsymbol{Y}}(\boldsymbol{x},\boldsymbol{y})}{p_{\boldsymbol{X}}(\boldsymbol{x})} \quad\mathrm{and}\quad p_{\boldsymbol{X}}(\boldsymbol{x}) = \int\limits_{\forall \boldsymbol{y}} p_{\boldsymbol{X},\boldsymbol{Y}}(\boldsymbol{x},\boldsymbol{y}) \mathrm{d}\boldsymbol{y},
\label{DAGMControl_f2}
\end{equation}
which replaces the $\argmax$-function with the expectation value, is often a good approximation.  

Usually, probability densities have to be estimated based on sample data $D = \{(\boldsymbol{x}_1,\boldsymbol{y}_1),\ldots,(\boldsymbol{x}_n,\boldsymbol{y}_n)\}$. One way is the non-parametric kernel density estimation method, which turns every data point into a center of a kernel function. A compression algorithm can reduce the high memory consumption of this method. The resulting model structure is
\begin{equation}
\tilde{p}_{\boldsymbol{X},\boldsymbol{Y}}(\boldsymbol{x},\boldsymbol{y}) = \sum\limits_{k=1}^m a_k \phi(\boldsymbol{x} - \overline{\boldsymbol{x}}_{k},\boldsymbol{s}_{xk}) \phi(\boldsymbol{y} - \overline{\boldsymbol{y}}_{k},\boldsymbol{s}_{yk}),
\label{DAGMControl_f3}
\end{equation}
which corresponds to a \textit{mixture model}. The $a_k$ are weights with a sum of one. The $\overline{\boldsymbol{x}}_{k}$ and $\overline{\boldsymbol{y}}_{k}$ are the centers of the kernels, while the $\boldsymbol{s}_{xk}$ and $\boldsymbol{s}_{yk}$ determine the smoothness of the estimation. An example for the kernel function $\phi$ is a Gaussian 
with diagonal covariance matrix.

Inserting (\ref{DAGMControl_f3}) into equation~(\ref{DAGMControl_f1}) and using the relation~(\ref{DAGMControl_f2}) results in an estimator for $\boldsymbol{y_0}$, given $\boldsymbol{x_0}$:
\begin{equation}
\boldsymbol{y}_0 \approx \tilde{\boldsymbol{y}}_0 = \frac{\sum\limits_{k=1}^m \,a_k\, \overline{\boldsymbol{y}}_{k}\, \phi(\boldsymbol{x} - \overline{\boldsymbol{x}}_{k},\boldsymbol{s}_{xk})}{\sum\limits_{k=1}^m a_k\, \phi(\boldsymbol{x} - \overline{\boldsymbol{x}}_{k},\boldsymbol{s}_{xk})}
\label{DAGMControl_f5}.
\end{equation}
This is known as a Nadaraya-Watson estimator, named after its inventors \cite{Nadaraya64} and \cite{Watson64}. Usually, this method is of little practical value, because it interpolates poorly\footnote{Equidistant sample points are an exception: The Shannon interpolation theorem is a Nadaraya-Watson estimator and optimal, if the conditions of the sampling theorem are fulfilled.}. An improvement results from a combination with the standard least-mean-square method \citep{Cleveland79, Haerdle90, Wang90}. For this, the error criterion 
\begin{equation}
E_i = \sum\limits_{k=1}^{m} a_k \phi(\boldsymbol{x} - \overline{\boldsymbol{x}}_{k},\boldsymbol{s}_{xk})(\alpha_i + \boldsymbol{\beta}_i^T\overline{\boldsymbol{x}}_{k} - \overline{y}_{ki})^2
\label{DAGMControl_f6}
\end{equation}
is defined for all $i=1,\ldots,d_y$. A minimization regarding $\alpha_i \in \mathbb{R}$ and $\boldsymbol{\beta}_i \in \mathbb{R}^{d_x}$ delivers the regression function 
\begin{equation}
\tilde{y}_i = \hat{\alpha}_i + \hat{\boldsymbol{\beta}}_i^T \boldsymbol{x}
\label{DAGMControl_f7}.
\end{equation}
Note that this is not a common linear function, because the optima $\hat{\alpha}_i$ and $\hat{\boldsymbol{\beta}}_i$ depend on $\boldsymbol{x}$. Only for homogenous $a_k$ and very large values for the smoothing parameters $\boldsymbol{s}_{xk}$, the formula~(\ref{DAGMControl_f7}) becomes a usual linear regression, because the density estimation converges in this case against a uniform distribution. 

Note also that expression~(\ref{DAGMControl_f7}) is identical to the Nadaraya-Watson regression~(\ref{DAGMControl_f5}), if $\boldsymbol{\beta}_i$ is determined as zero by the minimization. This shows that the Nadaraya-Watson regression is only an estimator of the order zero. The first order estimator~(\ref{DAGMControl_f7}) usually has a significantly better quality. 

\subsection{The Associate Framework}

The Nadaraya-Watson approximation, especially the linear localizing version, is an efficient and powerful method for extracting information that is represented by a probability density of the form~(\ref{DAGMControl_f3}). A description of the density estimation approach that we have used for our experiments is beyond the scope of this work. We only want to mention here that we apply for the computation of the smoothing parameter an extended version of the method described by \cite{Duin76} with online, real-time, and compression capabilities. Furthermore, the approach is able to forget knowledge when it was not needed for a long time. 

\begin{figure}[th]
\centerline{\includegraphics[width=\textwidth]{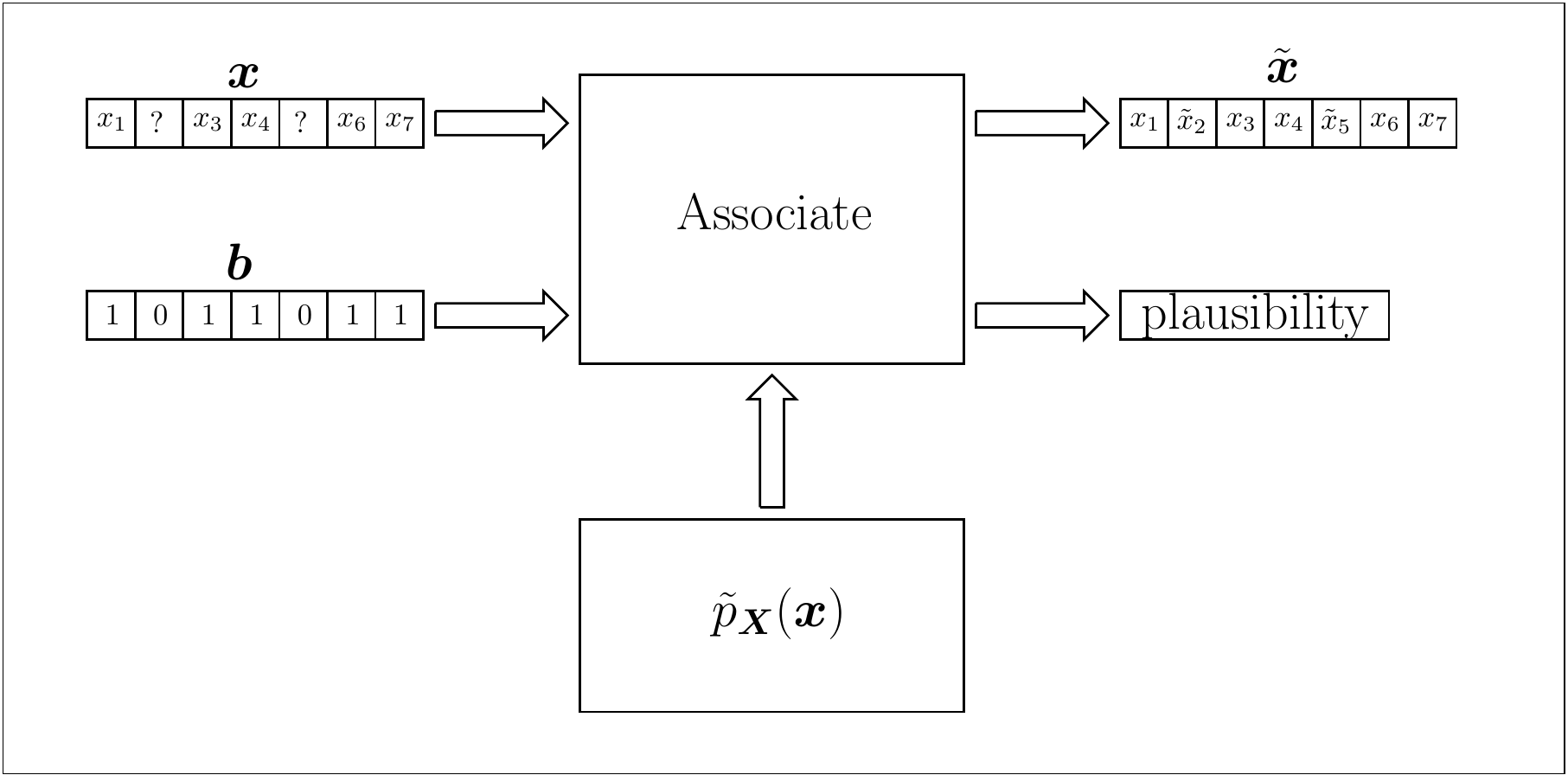}}
\caption{The function Associate may be used for regression, classification, and recognition. It estimates the missing components of the vector $\boldsymbol{x}$ and delivers a plausibility measurement for the reconstructed vector.}
\label{class_fig20}
\end{figure}

The framework, which is composed of density estimation and the Nadaraya-Watson approach, can be used for various pattern recognition applications, such as classification, recognition, reconstruction, and regression. Figure~\ref{class_fig20} shows the fundamental structure. In the center is the function Associate, which completes the missing elements in vector $\boldsymbol{x}$. The information, which elements are missing, is stored in the binary valued vector $\boldsymbol{b}$ with the same dimension. If a component is zero, the related component in vector $\boldsymbol{x}$ is missing and to be reconstructed. Furthermore, the function Associate assesses the quality of the reconstructed vector $\tilde{\boldsymbol{x}}$. This plausibility measurement can be interpreted as the estimated probability for even more unlikely vectors than $\tilde{\boldsymbol{x}}$. A detailed description of this criterion is given by \cite{Kuehn08}. 

The advantage of the framework is that data storing and data processing are entirely separated from each other. The density estimation works in parallel, independently, and fully automated. Its only purpose is to save the observed data, without adding or deleting information. The evaluation by the Associate function does not influence this knowledge-saving process. Hence, the framework is highly similar to a database: the probability density contains the knowledge and the Associate function makes it available.

We will illustrate this with a short, theoretical example. Let us assume that we have an arbitrary classification problem, which we want to solve. Usually, a training dataset is given
\begin{equation}
D = \{(\boldsymbol{m}_1,\boldsymbol{c}_1),\ldots,(\boldsymbol{m}_n,\boldsymbol{c}_n)\}.
\end{equation}
The $\boldsymbol{m}_i$, with $i \in [1,n]$, are pattern vectors with dimension $d_m$ and the $\boldsymbol{c}_i$ are class information vectors that encode the class to which $\boldsymbol{m}_i$ belongs. The dimension of such a vector corresponds to the number of classes $d_c$, with the $j$th component containing a measurement between zero and one for the affiliation of the pattern vector $\boldsymbol{m}_i$ to the $j$th class.

In order to use the associate framework, it is necessary to combine the vectors $\boldsymbol{m}_i$ and $\boldsymbol{c}_i$ to vectors $\boldsymbol{x}_i = (\boldsymbol{m}_i,\boldsymbol{c}_i) \in \mathbb{R}^{d_m + d_c}$. After that, the probability density $p_{\boldsymbol{X}}(\boldsymbol{x})$ of the random variable $\boldsymbol{X}$ can be estimated based on the training dataset $D = \{\boldsymbol{x}_1,\ldots,\boldsymbol{x}_n\}$. The Associate function can now classify new pattern vectors $\boldsymbol{m}'$ by passing the vectors $\boldsymbol{x} = (\boldsymbol{m}_1',\ldots,\boldsymbol{m}_{d_m}',?,\ldots,?)$ and $\boldsymbol{b} = (1,\ldots,1,0,\ldots,0)$ to the function. The reconstructed elements in the resulting vector $\tilde{\boldsymbol{x}}$ correspond to the class information vector. Furthermore, the function delivers the plausibility of the decision. This measurement is important if the class number is zero. In this case, there is no class information vector to reconstruct and the decision by the plausibility becomes a pure recognition. Note that it is possible to show that the classification by this framework is identical to a Bayes-classification.

Furthermore, the framework offers some additional possibilities. For example, it is possible to mark an existing element of the pattern vector $\boldsymbol{m}_0$ as unknown. This induces the Associate function to reconstruct this element as well. If this leads to a significant rising of the plausibility, this can be considered as an indication that the original value in the pattern vector was corrupt. A careful utilization of this feature enables a classification even in cases were parts of the pattern vectors are corrupted.

But the application of the framework is not limited to classification or recognition. The most interesting application is regression. The following section will show this using a practical example.

\section{Modeling and Control of a Motorcar Powertrain}

\subsection{Modeling}

First, we will describe the modeling of a motorcar powertrain with automatic gearbox in normal mode, before we show how the model can be used for control engineering. For reasons of convenience and because there is no reason for the assumption that the modeling would not be successfully for a real car, we will use here instead a complex Modelica\trademark/Dymola\trademark~model, which physically models engine, gearbox, and the other components. 

\begin{figure}[th]
\centerline{\includegraphics[width=\textwidth]{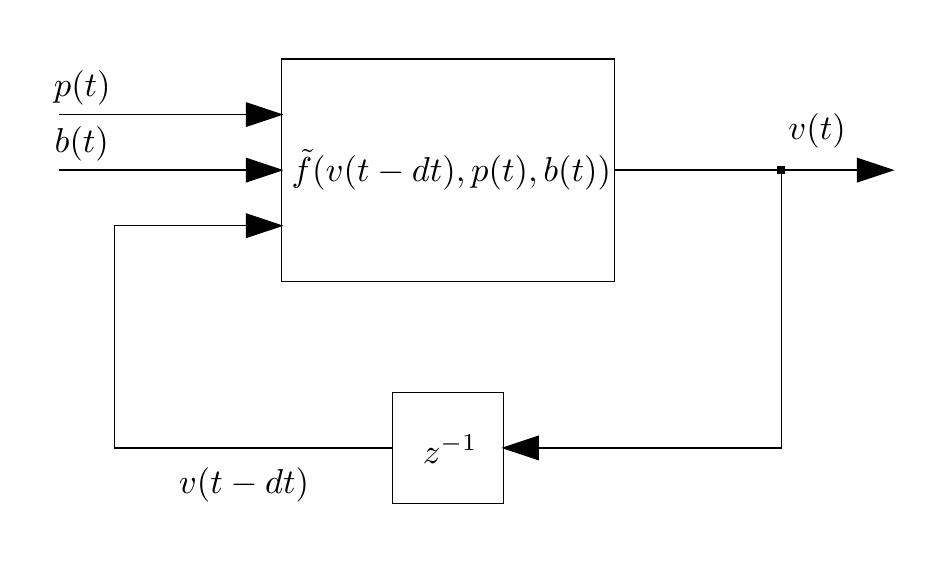}}
\caption{The model structure for the description of the dependencies between accelerator pedal position $p(t)$, brake pedal position $b(t)$, and car speed $v(t)$. The element $z^{-1}$ delays its input signal by a time unit $dt$.}
\label{class_fig33}
\end{figure}

Essentially, it is possible to describe the speed $v(t)$ of a car for the point in time $t$ as a function of the accelerator pedal position $p(t) \in [0\ldots1]$, the brake pedal position $b(t) \in [0\ldots1]$, and the speed of the car for the point in time $t-dt$, with $dt$ denoting a small time unit. Figure~\ref{class_fig33} shows the structure of the model. The feedback by the delay element is necessary, because the current speed $v(t)$ depends to a high degree on the speed a moment before. Note that this model structure is only an approximation, which may be sufficient or not. Technical systems usually require further feedback loops with different time delays.

After the model structure definition, an estimation of the function $\tilde{f}$ is possible. According to the associate framework, the inputs and outputs per time step are combined to vectors
\begin{equation}
\boldsymbol{x}(t) = (v(t-dt),v(t),p(t),b(t)).
\end{equation}
With a set of such vectors, a modeling or estimation of the probability density $p_{\boldsymbol{X}}(\boldsymbol{x})$ is possible. There are two ways to get an appropriate sequence of vectors $\boldsymbol{x}(t)$. The first, easier, option is to observe a driver. The second is to try various inputs and to study the reactions of the systems

\begin{figure}[th]
\centerline{\includegraphics[width=\textwidth]{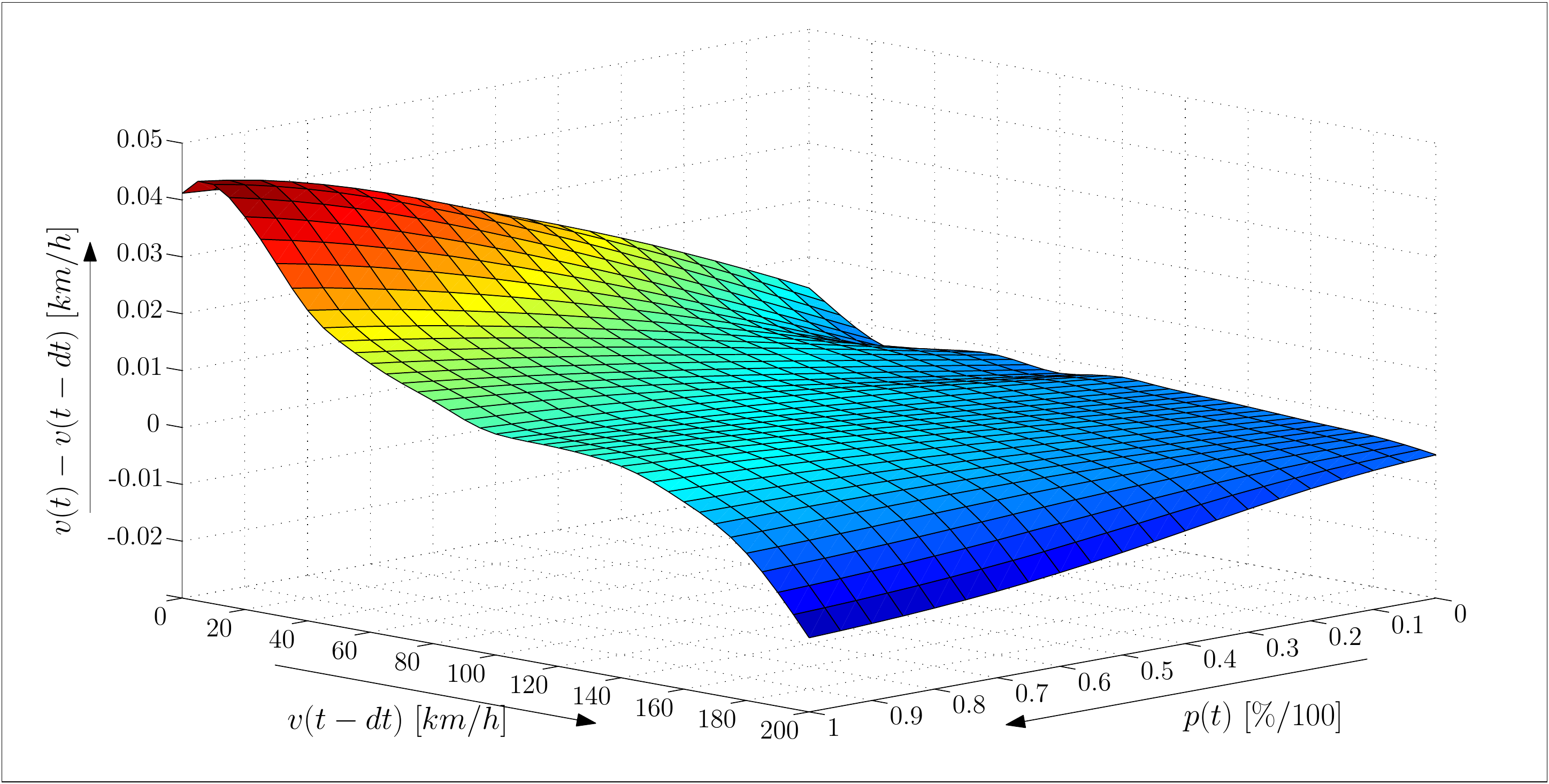}}
\caption{The results of the modeling. The speed $v(t)$ is a prognosis based on the acceleration pedal position $p(t)$ and the previous speed $v(t-dt)$. The brake pedal position was disregarded and assumed as zero for this representation.}
\label{class_fig30}
\end{figure}

Figure~\ref{class_fig30} shows the regression function that results from the online modeling with our framework. It includes the behavior of the Modelica\trademark~powertrain model for $500$ seconds with a time resolution of one millisecond. The high time resolution was a requirement of the Modelica\trademark~model and could be most likely reduced for a real car. We have roughly approximated the driver by a random walk process.

Because of the simplifying model structure, it is necessary to consider large amounts of sample points in order to average out the dependencies of internal states. A modeling with a spline based upon some pre-selected sample points would not lead to this result. Hence, figure~\ref{class_fig30} does not show a precise relation, but describes only the \textit{most probable} reaction of the powertrain\footnote{However, the regression function itself could be used as a basis for a spline model.}. 

The speeds $v(t)$ and $v(t+dt)$ differ only little from each other. For this reason, the graph of the estimated function $\tilde{f}(v(t-dt),p(t),b(t))$ would result in a simple slant, making the essential information difficult to recognize. Hence, figure~\ref{class_fig30} shows, instead of $\tilde{f}(v(t-dt),p(t),b(t))$, the difference $\tilde{f}(v(t-dt),p(t),b(t)) - v(t-dt)$, which could be interpreted as acceleration. For instance, a fully applied acceleration pedal leads to the highest acceleration at a low speed. On the other hand, the car hardly accelerates when the engine is idle and slows down noticeably for speeds over $60~km/h$. The highest speed of the simulated car can be estimated to be approximately $180~km/h$.

\subsection{Feedback Control \label{Selbstlernender Regler}}

It is only a small step from modeling to control. If it is known how a system will react to an input, it is also possible to manipulate it systematically. For this, however, it is necessary to invert the model, because in its original form it can only predict the reaction to a given input. For a precise manipulation, on the contrary, it is necessary to find an input that could result in the desired output. This means that the output is given and we need to find an appropriate input.   

This is not a problem for our framework, because the Associate function (see figure~\ref{class_fig20}) is able to use any component as input or output. The density estimation delivers predictions in both directions, because input and output are modeled in the same way by a common distribution. In the following, we will demonstrate this using the powertrain example.

The sample problem is to accelerate and decelerate the car to attain a given target speed. Because the target speed can be changed discontinuously and the car cannot reach arbitrary accelerations due to its inertia, not all target speed graphs can be realized. But it is desired that the car reaches the target speed as fast as technically possible.

\begin{figure}[th]
\centerline{\includegraphics[width=\textwidth]{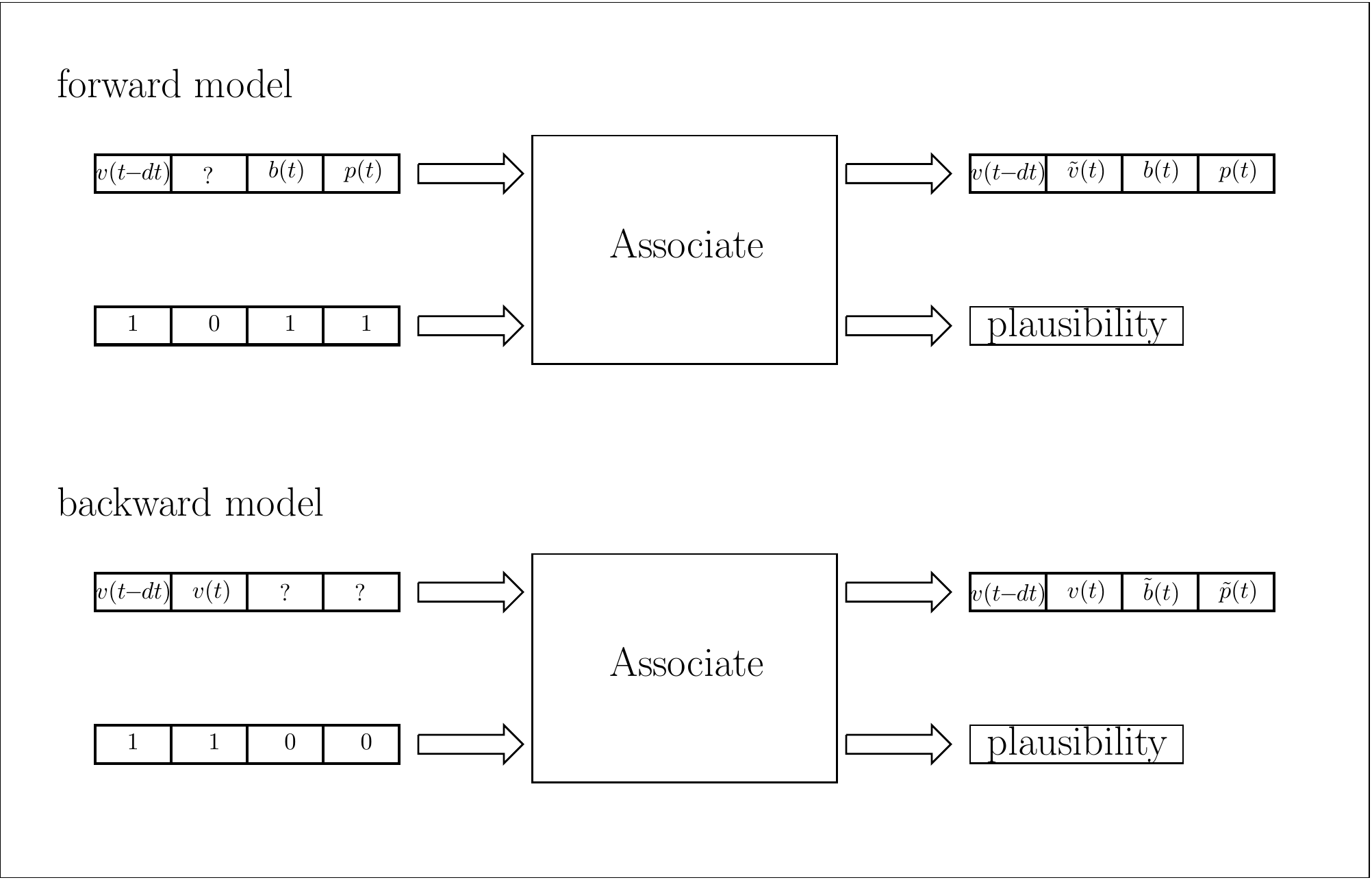}}
\caption{The associate function enables a forward and a backward prediction.}
\label{class_fig37}
\end{figure}  

The estimated probability density $\tilde{p}_{\boldsymbol{X}}(\boldsymbol{x})$ and the Associate function allow us both to predict the behavior of the motorcar by a forward model and to guess the $b(t)$ and $p(t)$ if $v(t)$ is given and the current state of the car $v(t-dt)$ is known.
Figure~\ref{class_fig37} shows the application of the Associate function for the implementation of \textit{forward} and \textit{backward} prediction.

The backward model is also suitable as feedback controller, because a feedback controller computes for reference and current values appropriate inputs. The well-known PID~controller computes these inputs from the difference between setpoint and process variable using digital filters or equivalent algorithms. This approach is fundamentally different from the method described here.

There are several important differences. One reason is that the models used here are non-linear to a high degree, another one that the Associate function does not compute an error, that is, a difference between setpoint and process variable. Instead, it guesses which inputs \textit{could} result in the desired setpoints. These will then be applied to the controlled system, which leads to a reduction of the error, if the knowledge was sufficient. Over the next steps, the error becomes smaller and smaller, until the current value reaches more or less the setpoint. At the same time, the controller can improve its knowledge about the controlled system, because the applied input leads to a reaction that can be evaluated, regardless of whether it was the desired reaction or not. Both together, action and reaction, can be used as further examples for the estimation of the probability density.

Small, random deviations of the controlled system from the 'normal behavior' cannot be foreseen by the controller, because it has no knowledge about the influences causing these disturbances. If it would have this knowledge, the behavior would be no longer stochastic, but deterministic and predictable. In practice, there are always disturbances triggering reactions of the system that do not correspond to the exact expected behavior. Because actions and reactions are modeled by a probability density, the controller essentially knows this degree of uncertainty. This allows it to express its expectations in advance in the form of a plausibility criterion.

Sometimes, the normal behavior will change over time. Because of the permanently running update procedure of the estimated probability density, the controller accommodates these changes automatically. This means that the controller is able to learn and to compensate varying environmental conditions. A special case of this auto-accommodation scenario is where the controller starts operation without having an estimated density. The only thing the controller can do at first is to guess. After setting the guessed values, the controller acquires some knowledge about the system from the reaction of the system it is controlling. This information can be used for the modeling of the density estimation. In the next step, the controller has to guess again. But again, its knowledge improves. After many good or bad decisions, the estimated probability density is good enough and the decisions of the controller are no longer blind guesses, but chosen and precise actions.

This self-contained way to learn by trial and error is risky of course, because it can result in damages to the controlled machine. But it is also possible to implement the initial learning by a 'teacher', who demonstrates a reasonable behavior. For our example, this teacher could be a human driver. The controller only has to observe him until its predictions are good enough. After this, the backward model can replace the driver.

\begin{figure}[th]
\centerline{
\includegraphics[width=0.5\textwidth]{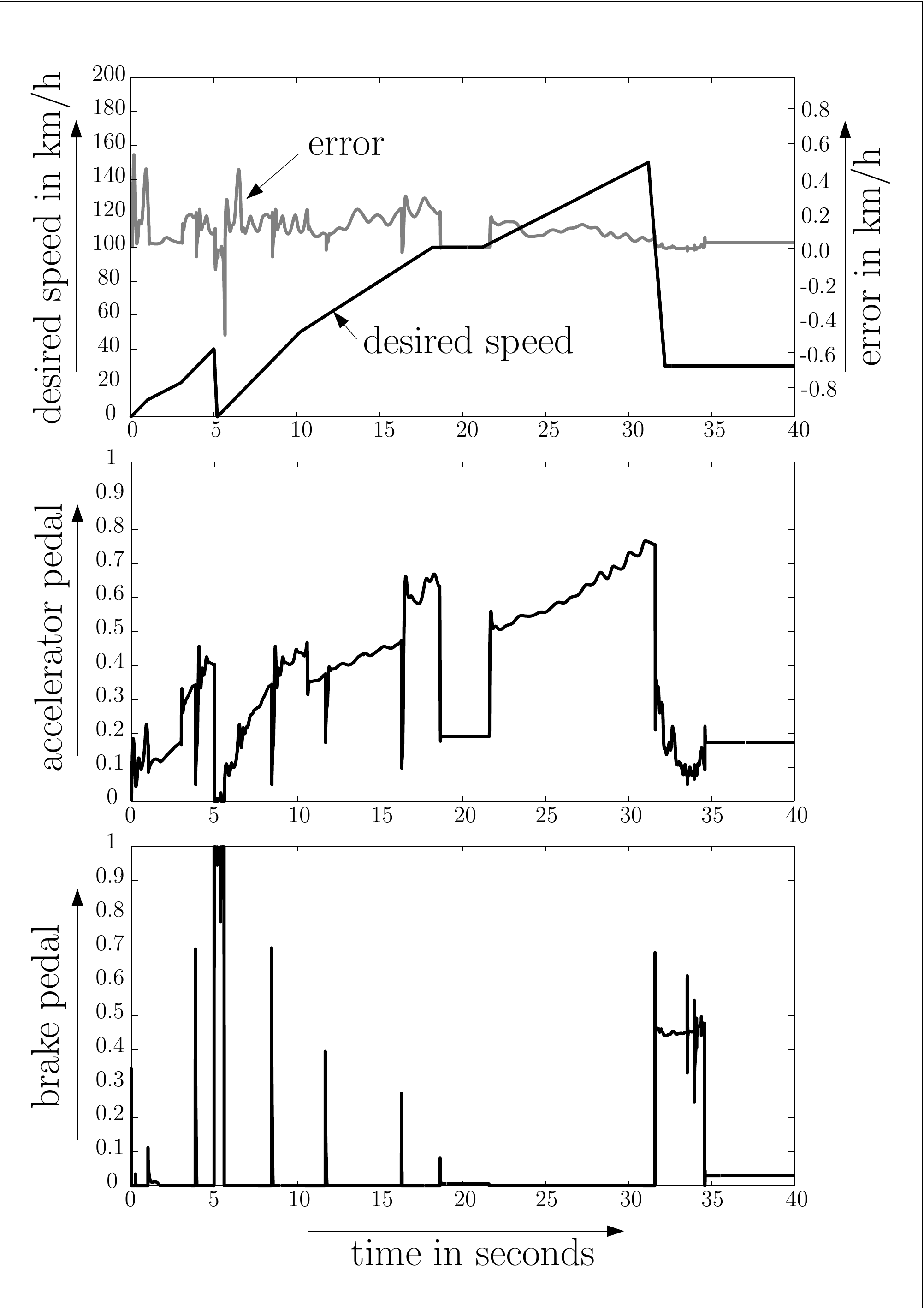}
\includegraphics[width=0.5\textwidth]{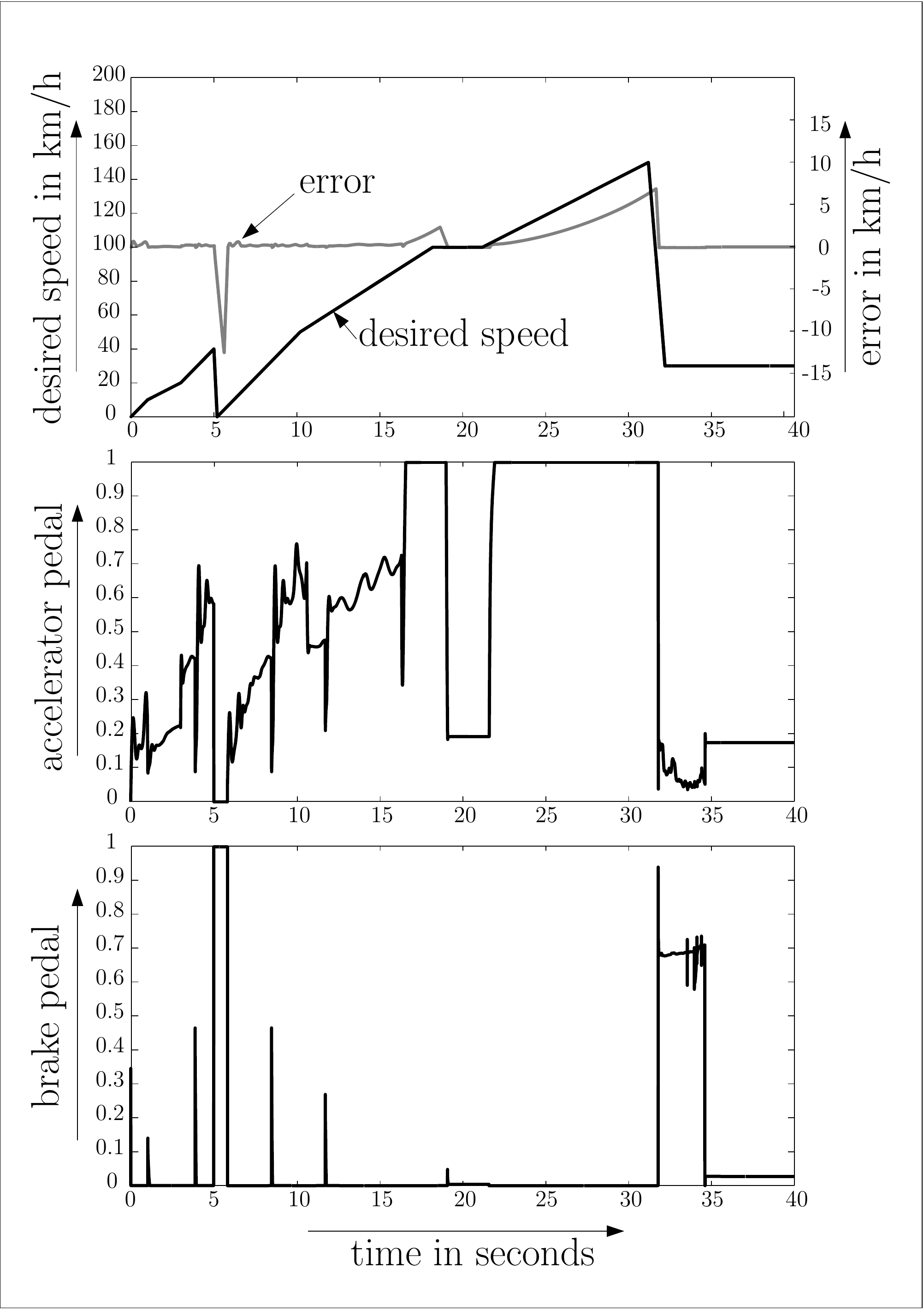}
}
\caption{The results of our controller experiment: The left plot shows that the controller achieves the desired speed nearly perfectly. The error is everywhere less than $1\,km/h$. The right side shows the results of increasing the car mass from $1800\,kg$ to $2800\,kg$. The controller still works, the deviations are only the result of physical limitations.} 
\label{class_fig38}
\end{figure}  

We will now complete these theoretical considerations with a discussion of the practical results of our powertrain example. The initial model was the estimated probability density of the last section. What is most noticeable when using the backward model is that the density has only non-zero values for $v(t-dt) \approx v(t)$. The reason is that within the short time period $dt$ no large speed changes occurred, something that is obviously technically impossible. Accordingly, the density was small if, for instance, the current speed $v(t-dt)$ was zero, but a speed of $v(t) = 100\,km/h$ was desired. Nevertheless, the Associate function was immediately able to deliver good extrapolations. In this case, the controller would have made the decision to open the accelerator pedal far beyond the physical end-point. Although this is not possible, it is correct in the tendency. For this reason, all values for brake and accelerator were limited to values in the valid range between zero and one. This would allow to immediately realize a functioning control.

For the simulation, the controller was tested with different, arbitrarily chosen speed demands. An example is presented in figure~\ref{class_fig38}. The left plot shows that the speed attained corresponds almost exactly to the desired speed. Also, a sudden increasing of the car mass from $1800\,kg$ to $2800\,kg$ does not lead to problems, despite that no longer every desired speed is realizable. For instance, the engine is for speeds over $100\,km/h$ simply not powerful enough, although the accelerator pedal is opened completely. Furthermore, the car cannot brake fast enough sometimes. This control error cannot be avoided.

\section{Conclusion}

A motorcar powertrain is a complex technical system and its physical modeling is difficult. But a black box modeling with the associate framework is simple. Already after less than 10 minutes observation, the associate framework has a detailed model of the system and is able to control it. This seems to be not least due to the fact that the system can be described very well by only four input and output values. Many internal details, which have to be considered by a physical modeling, can be neglected, because they are not visible from the outside.

The associate framework obviously imitates a human driver. Even a professional driver does not have always a detailed physical idea of how the engine inside his car works. Nevertheless, he is able to control his car. He learns and improves this ability by observation and permanent use. Just like a human, the associate framework has no need for an initial set of parameters or external interventions by a supervisor. This becomes possible by a strict separation of modeling and evaluation. The non-parametric density estimation, which runs permanently in the background, has the only purpose to save the observed knowledge as detailed and undistorted as possible. The evaluation of this information, for example, for the prediction, is based solely on the laws of probability theory. 



\begin{thebibliography}{7}

\bibitem[{Cleveland(1979)}]{Cleveland79}
W.S. Cleveland.
\newblock Robust locally weighted regression and smoothing scatterplots.
\newblock {\em Journal of the American Stastical Association}, 1979.

\bibitem[{Duin(1976)}]{Duin76}
R.P.W. Duin.
\newblock On the choice of the smoothing parameters for parzen estimators of
  probability density functions.
\newblock {\em IEEE Transactions on Computers}, Vol. C-25, No. 11:1175--1179,
  1976.

\bibitem[H\"ardle(1990)]{Haerdle90}
W.~Härdle.
\newblock {\em Applied Nonparametric Regression}.
\newblock Cambridge University Press, 1990.

\bibitem[K\"uhn(2008)]{Kuehn08}
S.~K\"uhn.
\newblock Generalized prediction intervals for arbitrary distributed
  high-dimensional data.
\newblock {\em ArXiv.org:0809.3352}, 2008.

\bibitem[{Nadaraya(1964)}]{Nadaraya64}
E.~A. Nadaraya.
\newblock On estimating regression.
\newblock {\em Theory of Probability and its Applications}, Vol. 9:141--142,
  1964.

\bibitem[Wang(1990)]{Wang90}
F.T. Wang.
\newblock {\em A New Method for Robust Nonparametric Regression}.
\newblock PhD thesis, Rice University, 1990.

\bibitem[Watson(1964)]{Watson64}
G.~S. Watson.
\newblock Smooth regression analysis.
\newblock {\em Sankhya, Series A}, 26:359--372, 1964.

\end{thebibliography}
\end{document}